\documentclass[10pt,twocolumn,letterpaper]{article}

\usepackage[final]{paper}  
\usepackage{graphicx}
\usepackage{amsmath}
\usepackage{amssymb}
\usepackage{booktabs}
\usepackage{multirow}  
\usepackage{bbding}

\usepackage[table]{xcolor}
\definecolor{mygray}{gray}{.92}

\makeatletter
\newcommand{\thickhline}{%
    \noalign {\ifnum 0=`}\fi \hrule height 1pt
    \futurelet \reserved@a \@xhline
}

\usepackage[pagebackref,breaklinks,colorlinks]{hyperref}

\usepackage[capitalize]{cleveref}
\crefname{section}{Sec.}{Secs.}
\Crefname{section}{Section}{Sections}
\Crefname{table}{Table}{Tables}
\crefname{table}{Tab.}{Tabs.}

\begin{document}

\title{TAL: Two-stream Adaptive Learning for Generalizable Person Re-identification}

% \and
\author{Yichao Yan,
Junjie Li,
Shengcai Liao,
Jie Qin,
Bingbing Ni,
and Xiaokang Yang
\\
% $^{3}$ MoE Key Lab of Artificial Intelligence, AI Institute, Shanghai Jiao Tong University, China \qquad \\
$^{}$ {\tt\small yanyichao@sjtu.edu.cn}
}
\maketitle

%%%%%%%%% ABSTRACT
\begin{abstract}
   Domain generalizable person re-identification aims to apply a trained model to unseen domains. Prior works either combine the data in all the training domains to capture domain-invariant features, or adopt a mixture of experts to investigate domain-specific information. In this work, we argue that both domain-specific and domain-invariant features are crucial for improving the generalization ability of re-id models. To this end, we design a novel framework, which we name two-stream adaptive learning (TAL), to simultaneously model these two kinds of information. Specifically, a domain-specific stream is proposed to capture training domain statistics with batch normalization (BN) parameters, while an adaptive matching layer is designed to dynamically aggregate domain-level information.
   In the meantime, we design an adaptive BN layer in the domain-invariant stream, to approximate the statistics of various unseen domains. These two streams work adaptively and collaboratively to learn generalizable re-id features. Our framework can be applied to both single-source and multi-source domain generalization tasks, where experimental results show that our framework notably outperforms the state-of-the-art methods.

\end{abstract}

%%%%%%%%% BODY TEXT
\section{Introduction}
\label{sec:intro}

Person re-identification (re-id)~\cite{DBLP:journals/corr/abs-2001-04193} has been actively studied in the computer vision community. Over the past few years, fully supervised re-id has achieved remarkable progress, together with the rapid development of deep learning. In the supervised setting, re-id models are typically trained and evaluated within the same domain. Consequently, their performance may be significantly degraded when applied to unseen target domains due to the large discrepancy between the source/training and target/test domains. In other words, their generalization abilities are quite limited, preventing them from being adapted to various practical applications.
%However, in real-world applications, it is desirable that the models not only work well in the training domain, but can also handle various target domains. 
To address this, many unsupervised domain adaptation (UDA) models have been proposed to bridge the domain gap~\cite{deng2018image,chen2019instance,song2020unsupervised,fu2019self}. In this regime, re-id models are first trained in the labeled source domain, and further fine-tuned in the unlabeled target domain. Although these models significantly improve the performance under cross-dataset evaluation, an additional time-consuming adaptation step is required, making them inefficient for real-world deployment. Moreover, target domains are not always available for adaptation, hindering the successful applications of UDA models. To circumvent the above shortcomings, domain generalizable (DG) re-id~\cite{song2019generalizable,liao2020interpretable,dai2021generalizable,jin2020style,zhao2021learning}, which aims to improve the generalization ability of re-id models without domain adaptation, has been receiving more and more attention over the past years. The goal of DG re-id is to work out-of-the-box regardless of the scenarios of target domains, highly desired by real-world applications.

\begin{figure}
    \centering
    \includegraphics[width=\linewidth]{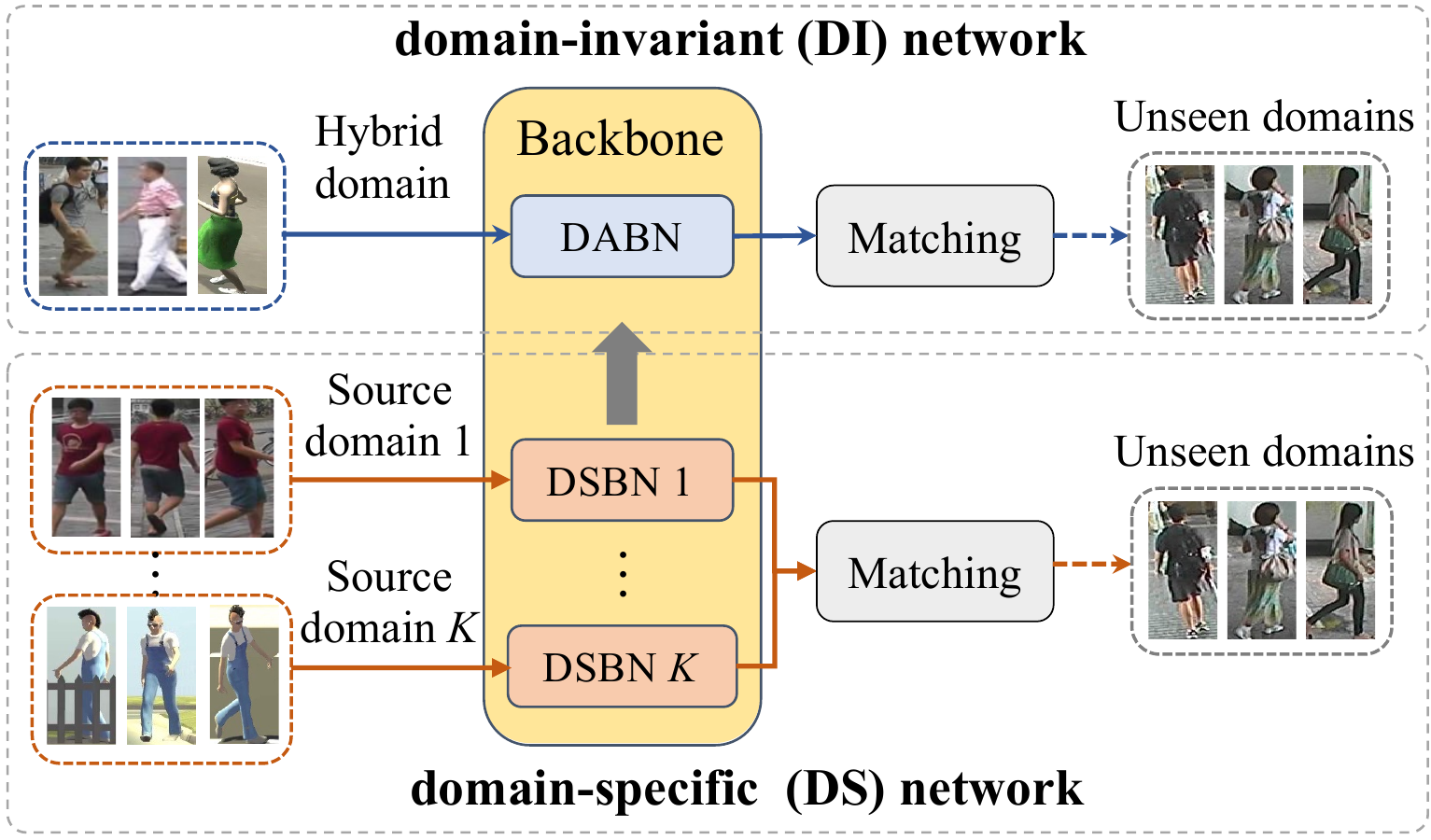}
    \caption{The proposed two-stream adaptive learning framework is composed of a domain-specific (DS) and a domain-invariant (DI) network. Both networks share the same backbone but with exclusive BN layers. In the DS network, DSBN captures domain-specific statistics for each domain expert. In the DI network, DABN adaptively approximates the unseen domain statistics from the DSBN parameters. These two networks simultaneously exploit the complementary DS and DI information and improve the generalization ability of the re-id model. }
    \label{fig:intro}
\end{figure}

%-------------------------------------------------------------------------

Previous efforts devoted to DG re-id can be generally divided into two categories. The first line of approaches, which we name single model frameworks, combine all the training data in the source domains and attempt to learn domain-invariant features via style normalization~\cite{jin2020style}, feature mapping~\cite{song2019generalizable} and meta-learning~\cite{DBLP:conf/cvpr/ChoiKJPK21}, \etc. However, the model parameters are fixed after training, and thus insufficient to be adapted to various target domains. More importantly, the \textbf{domain-specific information}, which can provide critical guidance to link the source and target domains, is ignored in these methods. The second line of works, which we refer to as the mixture of experts (MoE), develop a voting mechanism to aggregate the features from a series of domain-specific networks~\cite{dai2021generalizable}. However, the efficiency of these methods tends to decrease when there exist a large number of source domains. In the meantime, the \textbf{domain-invariant information} is overlooked with the disentangled treatment from each expert.

To address the above issues, we propose a two-stream adaptive learning (TAL) framework for DG re-id, to simultaneously capture these two types of information. As shown in Fig.~\ref{fig:intro}, our framework includes a domain-specific (DS) network that shares a similar spirit with the MoE framework, and a domain-invariant (DI) network that takes the hybrid source domain data as input. More specifically, on the one hand, we employ the domain-specific batch normalization (DSBN) layers~\cite{chang2019domain} in the DS stream to capture the statistics of the source domains, while other parameters in the backbone are shared. This not only makes the network more efficient in handling a large number of training domains, but also facilitates the domain-invariant feature learning in the convolutional layers. On the other hand, the DI network introduces a separate BN layer and is trained without domain labels from a hybrid dataset that contains all the training domains. In this way, both domain-specific and domain-invariant features can be simultaneously captured in our framework, providing complementary information to improve the generalization ability of our re-id model.

To fully unleash the potential of both streams and make our framework better generalize to the unseen domains, we introduce two adaptive learning modules. First, it is challenging for the fixed BN layers learned from the hybrid dataset to adapt to various unseen target domains. To this end, we design a novel domain-adaptive BN (DABN) layer in the DI stream, to dynamically approximate the target domain statistics from the domain-specific BN layers. Second, instead of explicit voting, we propose a domain-adaptive matching layer in the DS stream to better aggregate features from the experts. These two strategies mimic the unseen domain during training, and thus significantly improve the adaptability of our model when applied to novel domains.

In summary, our main contributions include:
\begin{itemize}
\setlength{\itemsep}{0pt}
	\setlength{\parsep}{-3pt}
	\setlength{\parskip}{-0pt}
	\setlength{\leftmargin}{-15pt}
% 	\vspace{-7pt}
    \item We propose a two-stream adaptive learning (TAL) framework for generalizable person re-id, which simultaneously captures the complementary domain-specific and domain-invariant information.

    \item We design two adaptive learning modules to approximate the unseen domain statistics, which significantly improve the adaptation and generalization abilities of our framework.
    
    \item Extensive experimental results on four large-scale datasets show that our framework outperforms state-of-the-art models in terms of both single-source and multi-source DG re-id tasks.
\end{itemize}

%------------------------------------------------------------------------
\section{Related Works}
\label{sec:related}

%-------------------------------------------------------------------------
\subsection{Person Re-identification}

Traditional person re-id methods mainly depend on hand-crafted features, such as~\cite{lowe2004distinctive,farenzena2010person}. With the development of deep learning, recent re-id models are typically based on deep neural networks, including but not limited to part-based feature representation learning~\cite{sun2018beyond,sun2019learning,miao2019pose}, distance metric learning~\cite{ahmed2015improved,DBLP:journals/corr/HermansBL17,chen2017beyond,chen2017person,sun2020circle}, and video-based learning~\cite{yan2020learning,zheng2016mars,wu2018exploit}. 
Despite their remarkable success, these models are fully-supervised. Due to domain shift, their performance will be significantly degraded if directly applied to a novel domain.
However, it is difficult and time-consuming to obtain the annotated data for each target domain, which makes it less practical for real-world applications. Therefore, unsupervised domain adaptation (UDA) re-id~\cite{song2020unsupervised,fu2019self,DBLP:conf/nips/Ge0C0L20,mekhazni2020unsupervised} and domain generalizable (DG) re-id~\cite{song2019generalizable,jia2019frustratingly,jin2020style,liao2020interpretable,zhao2021learning} have recently attracted growing interest in the community.

%-------------------------------------------------------------------------
\subsection{Domain Adaptation for Person Re-id}
Unsupervised domain adaptation (UDA) is the task of training a model on the source domain with labels and transferring it to the target domain under the unsupervised setting. UDA has been widely studied in recent years and was introduced into re-id task by \cite{ma2013domain}. Current UDA methods can be generally categorized into two branches: (1) those using generative adversarial networks (GAN)~\cite{DBLP:journals/corr/GoodfellowPMXWOCB14} to transfer the style of the source domain labeled images to the target domain \cite{deng2018image,chen2019instance,liu2020unity}, and (2) those adapting to the target domain with pseudo labels generated by clustering \cite{song2020unsupervised,fu2019self,DBLP:conf/nips/Ge0C0L20,zhai2020ad} or assigning  soft labels \cite{wang2020unsupervised}. However, UDA re-id models still rely on target domain images for adaptation, which are not always available. In contrast, DG re-id does not require the target domain data during training, which is more practical.

\begin{figure*}[t]
\centering
\subfloat[Data with domain labels are employed to update the DSBN parameters.\label{subfig-1-1}]{%
   \includegraphics[width=0.48\linewidth]{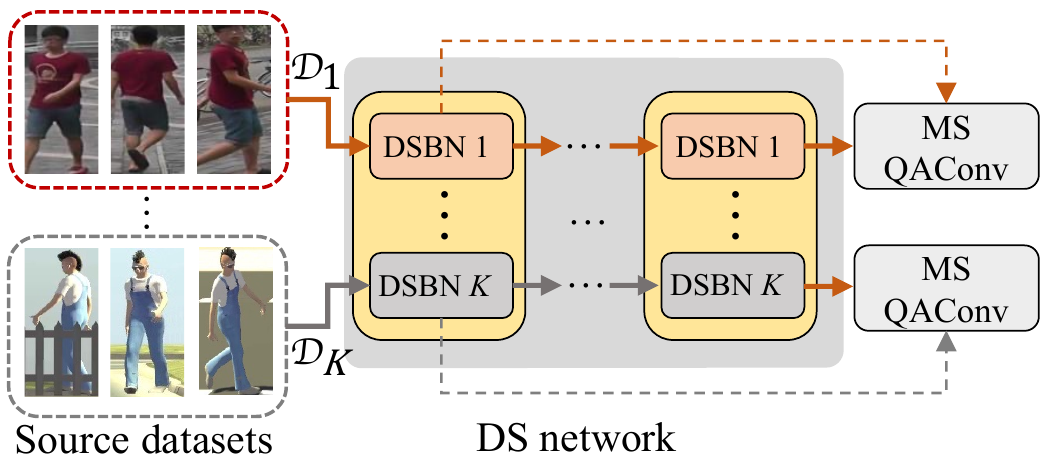}
}
\hspace{4mm}
%  \vspace{-2mm}
\subfloat[Data without domain labels are employed to update DABN layers. \label{subfig-1-2}]{%
   \includegraphics[width=0.48\linewidth]{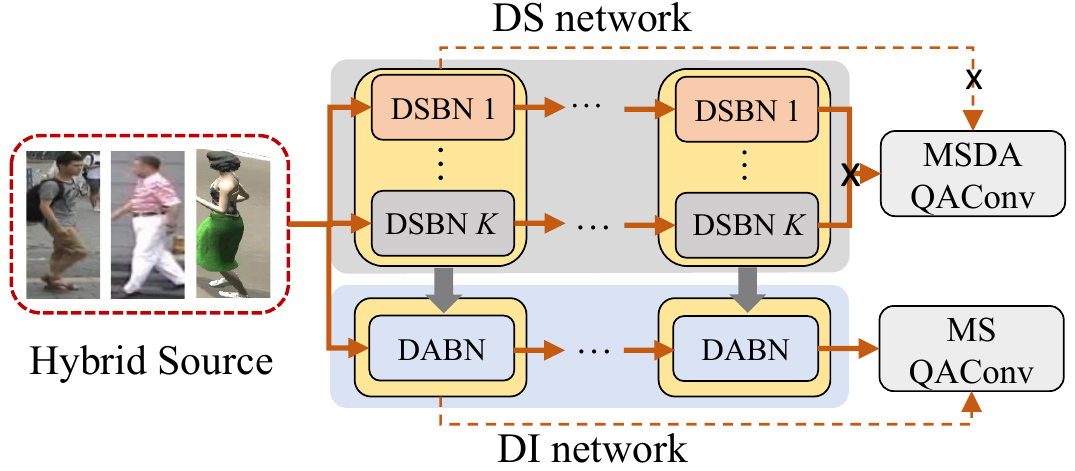}
}
\caption{Architecture of the proposed TAL framework. Both the DS and DI networks share the same backbone, but with different BN layers. The DSBN layers in the DS network are trained with data with domain labels, while the DABN layer in the DI network predicts the statistics of the unseen domain from DSBNs. `$\times$' denotes that the gradients do not flow back. }
 \label{fig:method}
%  \vspace{-6mm}
\end{figure*}

%-------------------------------------------------------------------------
\subsection{Generalizable Person Re-id}
The objective of domain generalization is to learn a robust, generalizable model, such that the learned model can obtain good performance on an unseen target domain without further updates. Existing DG re-id methods can be divided into two categories: (1) those that train a single model to learn domain-invariant features from single/multi-source domains \cite{song2019generalizable,liao2020interpretable,jin2020style,zhao2021learning}, and (2) those that employ a mixture of experts (MoE) to learn a series of separated experts for the subsets of whole source domains \cite{dai2021generalizable}. For the first category, 
%generalizable and robust feature representation learning is achieve by style normalization, domain invariant feature mapping and meta-learning methods. Specifically, 
Song \etal \cite{song2019generalizable} designed a Domain-Invariant Mapping Network (DIMN) for DG re-id. Jin \etal~\cite{jin2020style} proposed Style Normalization and Restitution (SNR) to disentangle the identity-relevant and identity-irrelevant features. Liao \etal~\cite{liao2020interpretable} proposed a query-adaptive matching mechanism to aggregate the local similarities for DG re-id. Zhao \etal~\cite{zhao2021learning} introduced a meta-learning strategy to simulate the training and test process of DG re-id. For the second category, MoE methods have been widely employed in scene parsing~\cite{fu2018moe} and image classification~\cite{ahmed2016network,wang2020deep}. Dai~\etal~\cite{dai2021generalizable} incorporated this strategy into the DG re-id task by designing a relevance-aware mixture of experts (RaMoE) that utilizes meta-learning for updating the voting network.
In this work, we take advantage of both approaches by introducing a two-stream framework to simultaneously exploit the domain-specific and domain-invariant information.

%-------------------------------------------------------------------------
\subsection{Normalization for Domain Generalization}
Recently, several works~\cite{pan2018two,DBLP:conf/nips/NamK18,chang2019domain,bai2021unsupervised,DBLP:conf/cvpr/ChoiKJPK21} have shown advantages of normalization techniques for boosting the generalization ability of neural networks. One of the most inspiring methods is domain-specific batch normalization (DSBN)~\cite{chang2019domain}, which employs exclusive BN layers to represent the statistics of the source and target domains. Nam \etal~\cite{DBLP:conf/nips/NamK18} designed a novel batch-instance normalization (BIN) layer to boost the collaboration of both BN and instance normalization (IN) layers. Choi \etal~\cite{DBLP:conf/cvpr/ChoiKJPK21} proposed a generalizable normalization layer by simulating unsuccessful generalization scenarios in the meta-learning pipeline. Inspired by these methods, we propose to employ the DSBN in our DS network, and utilize DSBN parameters to predict the statistics of the unseen domains in the DI network. In this way, both DS and DI information are exploited and thus yielding enhanced generalization ability.

%However, DSBN is not specifically designed for DG re-id problem. It may erase the distinctiveness of person features and cause performance drop in re-id task. Bai \etal \cite{bai2021unsupervised} proposed a rectification domain-specific batch normalization (RDSBN) module which simultaneously holds back person features and reduces domain-specific information.

\section{Methodology}
In this section, we introduce the proposed two-stream adaptive learning framework (\ie, TAL), for generalizable person re-id. We first elaborate on the proposed domain-specific and domain-invariant networks, and then present the training procedure of our framework.

\subsection{Domain-specific Network}
Suppose we have $K$ source domains $\{\mathcal{D}_1, ..., \mathcal{D}_K\}$ in the training set, where the $i$-th domain $\mathcal{D}_i = \{\mathbf{x}_i^1, ..., \mathbf{x}_i^{N_i}\}$ contains $N_i$ training images. Based on a shared backbone (\eg, ResNet-50~\cite{DBLP:conf/cvpr/HeZRS16}), the DS network contains a set of BN layers that are specifically engaged in each domain, as shown in Fig~\ref{subfig-1-1}. We denote the DS network as $\mathcal{F}_{DS} = \{\mathcal{F}_1, ..., \mathcal{F}_K \}$, which can be regarded as $K$ domain experts that share all the parameters except for the BN layers. Considering there are $P$ stages in the backbone, each domain expert will output a series of domain-specific feature maps:
\begin{equation}
   \mathbf{H}_i  = \mathcal{F}_i({\mathbf{x}_i}),
\end{equation}
where $\mathbf{H}_i =\{ \mathbf{h}_i^1,...,\mathbf{h}_i^P\}$, $\mathbf{x}_i \in \mathcal{D}_i$ and $i=1,...,K$. Subsequently, $\mathbf{H}_i$ are fed into a multi-scale variation of the query-adaptive convolutional layer~\cite{liao2020interpretable} (MS-QAConv), and supervised by a triplet matching loss $\mathcal{L}_{tri}$ (which will be introduced in Sec~\ref{sec:mtp}). 

As each domain expert in our DS network only receives single-source information, it is difficult for them to generalize to unseen domains. To address this, we employ a hybrid source to mimic the target domain, as shown in Fig.~\ref{subfig-1-2}, and again train the DS network without domain labels. Specifically, suppose $\mathcal{D}_{HS}$ contains all the training samples regardless of domain labels. Given input $\mathbf{x}_{HS} \in \mathcal{D}_{HS}$, we extract its features from all the domain experts:
\begin{equation}
    \mathbf{H}_{i,HS}  = \mathcal{F}_i({\mathbf{x}_{HS}}),
\end{equation}
where $i \in \{1,...,K\}$. The feature maps from all the domain experts $\{\mathbf{H}_{1,HS},...,\mathbf{H}_{K, HS}\}$ are aggregated by a domain-adaptive MS-QAConv layer, which we denote as MSDA-QAConv. The same matching loss $\mathcal{L}_{tri}$ is applied. Note that in this case, the gradients only flow back to the MSDA-QAConv layer, while the backbone parameters are no longer updated. In this way, the domain experts maintain the domain-specific information while being able to generalize to unseen domains.

\subsection{Domain-invariant Network}
\begin{figure}
    \centering
    \includegraphics[width=0.8\linewidth]{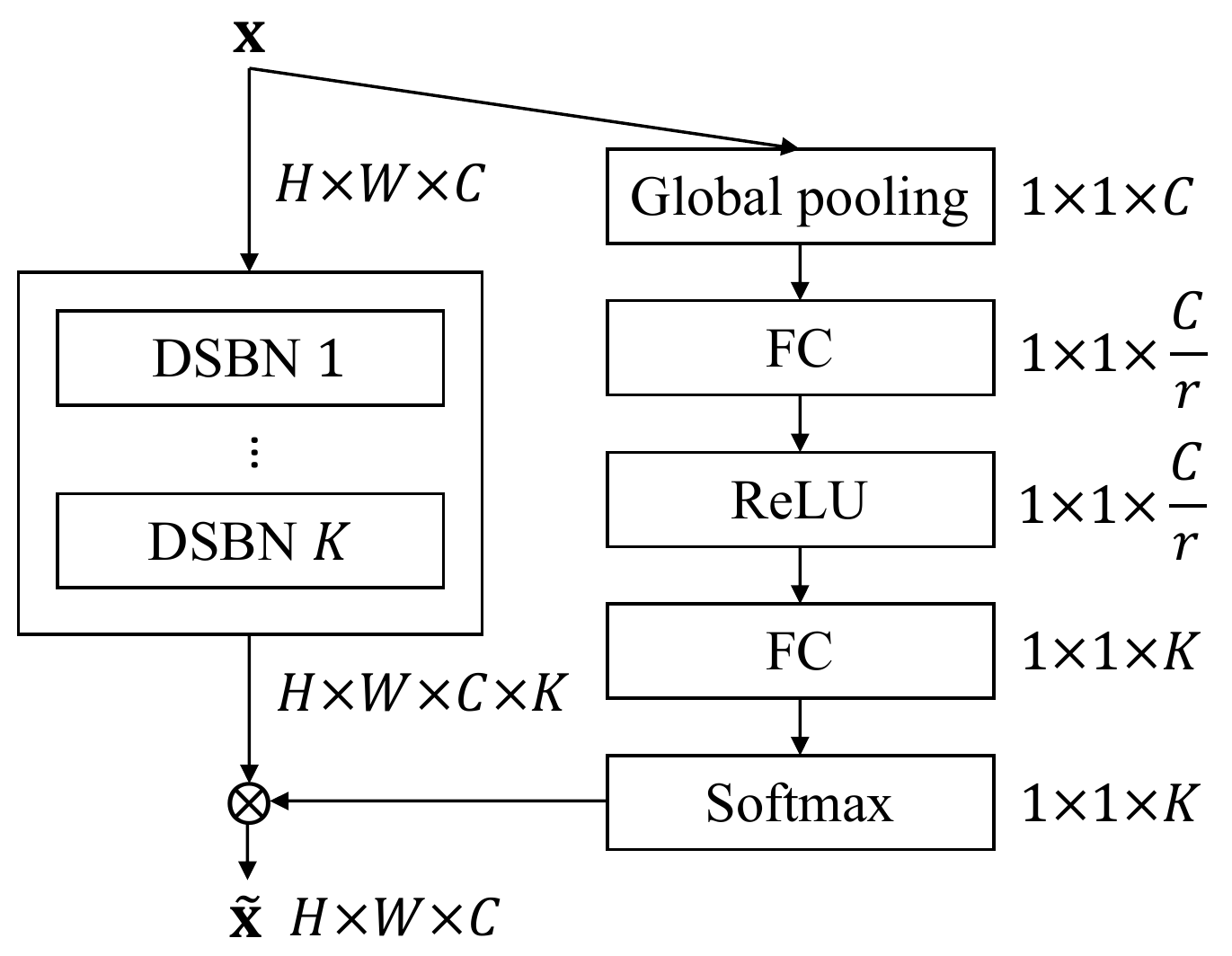}
    \caption{Illustration of the structure of DABN. The output of DABN is a weighted summation of the DSBN outputs, where the weights are calculated in a similar way as SENet~\cite{DBLP:conf/cvpr/HuSS18}.}
    \label{fig:dabn}
\end{figure}

Although the DS network tries to capture the domain-invariant information given the hybrid source, the feature aggregation only takes place at the last MSDA-QACovn layer, which is incapable of modeling the target domain statistics at different levels. To this end, we further introduce a DI network to normalize the features into domain-invariant representation. In particular, the DI network shares the same backbone with the DS network, but with specifically designed domain-adaptive BN (DABN) layers. To make the DABN layers adaptively approximate the target domain statistics, their normalization results are dynamically predicted from the DSBN layers, as shown in Fig.~\ref{subfig-1-2}. Suppose the input feature before the DABN layer is $\mathbf{x}$, we input $\mathbf{x}$ into each DSBN, where the output of the $i$-th domain can be represented as:
\begin{equation}
    {\rm DSBN}_i(\mathbf{x}, \gamma_i, \beta_i) = \gamma_i \hat{\mathbf{x}} + \beta_i, 
\end{equation}
where $\gamma_i$ and $\beta_i$ are the DSBN parameters and 
\begin{equation}
\hat{\mathbf{x}} = \frac{\mathbf{x} - \mu_i}{\sqrt{\sigma_i^2 + \epsilon}}.
\end{equation}
Here, $\mu_i$ and $\sigma_i$ are the mean and variance of the input feature.
In DABN, the normalization result $\Tilde{\mathbf{x}}$ can be calculated as a weighted summation of the DSBN outputs: 
\begin{equation} \label{eq:dabn}
    {\rm DABN}(\mathbf{x}, \alpha, \gamma, \beta) = \sum_{i=1}^{K}\alpha_i(\gamma_i \hat{\mathbf{x}} + \beta_i), 
\end{equation}
where the weighting terms $\alpha= \{\alpha_1, ..., \alpha_K\}$ are estimated in a similar way as squeeze-excitation network~\cite{DBLP:conf/cvpr/HuSS18} (SENet), as shown in Fig.~\ref{fig:dabn}. Specifically, the input feature maps are first input into an average pooling layer to extract global information. Then a bottleneck layer with ReLU activation is introduced to reduce the dimension (with reduction rate $r$), followed by another linear layer to predict the weights of each domain. The attention weights are normalized by a softmax layer:
\begin{equation} \label{eq:weight}
    \alpha = {\rm Softmax}(\mathbf{W}_2({\rm ReLU}(\mathbf{W}_1 {\rm GP} (\mathbf{x})))),
\end{equation}
where GP denotes global pooling operation, $\mathbf{W}_1$ and $\mathbf{W}_2$ are the parameters of the linear layers, respectively. $\alpha $ is applied to Eq.~\ref{eq:dabn} to calculate DABN activation.
% The detailed structure of DABN is shown in Fig.~\ref{fig:dabn}. 

\subsection{Model Training Procedure}\label{sec:mtp}
To make the model not only adaptive to novel domains, but also flexible to handle new test samples, we train our framework in the same fashion as QAConv~\cite{liao2020interpretable}. Specifically, QAConv tries to find local correspondences between two feature maps by constructing convolution kernels on-the-fly from one feature map, while performing convolution on the other feature map to calculate the local similarities. In our framework, we design two enhanced variations of QAConv, to further capture the multi-scale information (MS-QAConv), and to aggregate the domain-specific features (MSDA-QAConv).

\begin{figure}
    \centering
    \includegraphics[width=\linewidth]{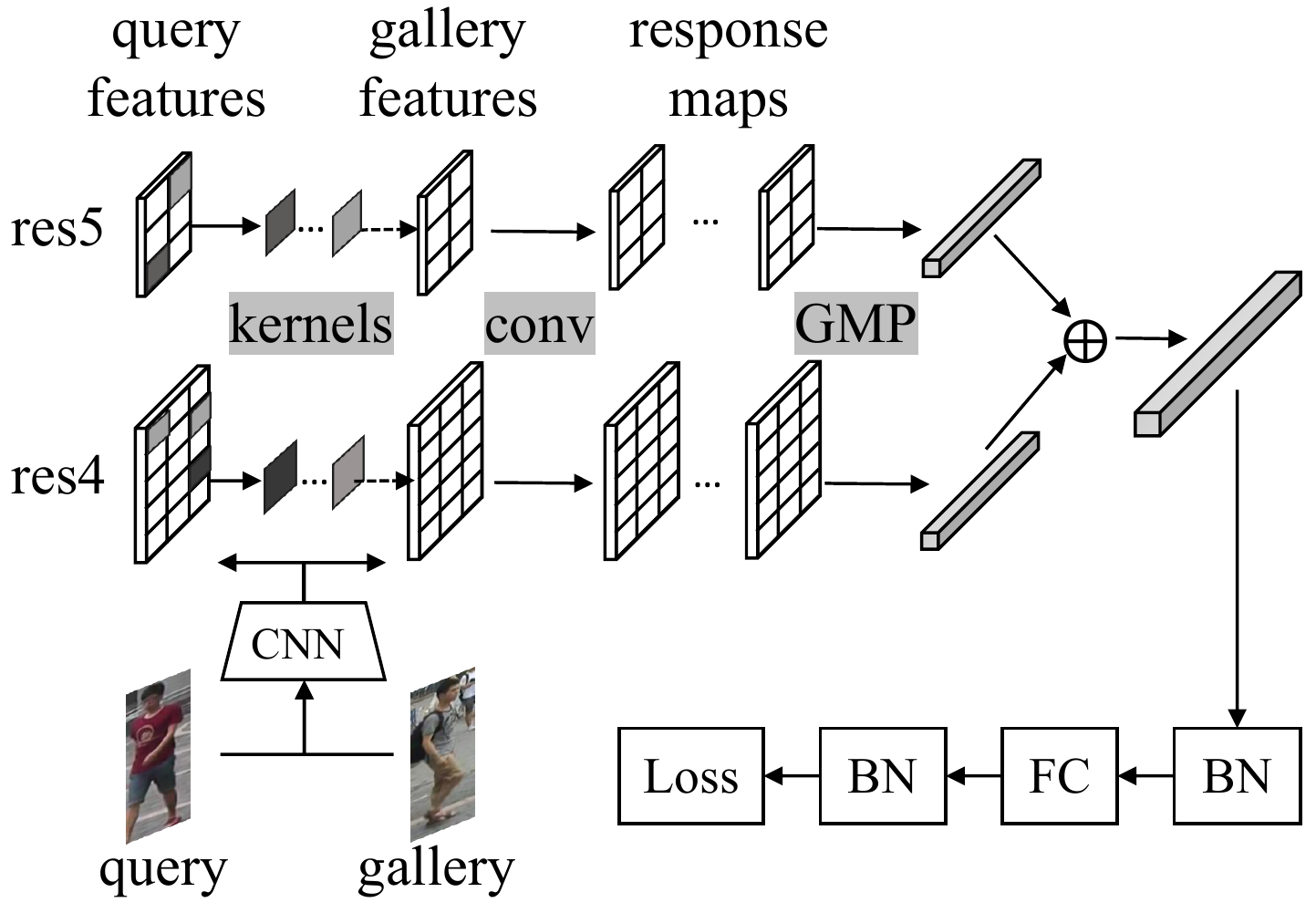}
    \caption{Illustration of MS-QAConv. Compared to QAConv~\cite{liao2020interpretable}, MS-QAConv computes multi-scale response maps and thus generates more reliable matching results by multi-scale fusion.}
    \label{fig:msqa}
\end{figure}

\textbf{MS-QAConv}.
The structure of MS-QAConv is illustrated in Fig.~\ref{fig:msqa}.
To match a pair of images, we first extract the multi-scale feature maps (\eg, from the res4 to res5 layers) of both the query and gallery images. Then the query feature maps are partitioned into local patches, which are re-organized into convolution kernels. By applying these kernels to the gallery feature maps, \ie, performing QAConv, we can get a set of response maps measuring the similarities between local patches. Then a global max pooling (GMP) operation is applied to these response maps to extract a feature vector that represents the best local correspondence. Vanilla QAConv directly take the vectors from a single scale (res4) into a BN-FC-BN block to calculate the final similarity value. However, as indicated by prior re-id works~\cite{DBLP:conf/iccvw/ChenZG17,DBLP:conf/mm/WangYCLZ18}, multi-scale feature representation contains hierarchical information to facilitate re-id learning. Therefore, MS-QAConv first aggregates the multi-scale matching information by concatenating the feature vectors from several scales, and then predicts the final similarity. In this way, the multi-level information is exploited to help our model make better predictions. 

\textbf{MSDA-QAConv}.
When an image without domain label is input into the DS network, it requires the QAConv block to be able to adaptively aggregate the response maps from different domain experts. Therefore, we further propose MSDA-QAConv to tackle this challenge. Specifically, the basic structure of MSDA-QAConv is similar to MS-QAConv, but it further contains an attention block to weight the response vectors from each domain. The attention weights are calculated in a similar way as in Eq.~\ref{eq:weight}, where the vectors are concatenated and fed into a FC-ReLU-FC-Softmax block to output the domain weights. Subsequently, the weighted summations of response vectors are calculated as the domain-level representation at each scale level, which are further aggregated to yield the final similarity. In this way, both multi-scale and multi-domain information are simultaneously captured for better adaptive feature learning.

\textbf{Loss Function}.
A triplet loss~\cite{DBLP:journals/corr/HermansBL17} with batch hard-negative mining is employed to supervise model training. Within a batch containing $B$ instances, we calculate the pairwise similarities between all the training samples. Specifically, we use $S_{i,p}$ to denote the similarities between the $i$-th sample and its positive pairs, while $S_{i,n}$ denotes the negative similarities. Then the batch hard triplet loss is calculated as follows:
\begin{equation}
\begin{aligned}
\mathcal{L}_{tri}= \sum_{i=1}^B [m -\min_{p}{S_{i,p}}+\max_{\substack{n}}{S_{i,n}}]_+ ,
\end{aligned}
\end{equation}
where $m$ is a hyperparameter denoting the margin between the positive and negative pairs, and $[*]_+=\max(*,0)$ ensures the loss value to be non-negative.

\textbf{Model Inference}. In the test phase, the output of the MSDA-QAConv block from the DS network and the output of the MS-QAConv from the DI network are combined to yield the final similarity.

\section{Experiments}

\subsection{Implementation Details}
Our implementation is based on the official code of QAConv\footnote{\hyperlink{}{https://github.com/ShengcaiLiao/QAConv}}~\cite{liao2020interpretable} and its improved variant (QAConv-GS~\cite{DBLP:journals/corr/abs-2104-01546}). We employ the IBN-Net50~\cite{pan2018two} pretrained on ImageNet~\cite{DBLP:conf/cvpr/DengDSLL009} as our default backbone. The features from res4 and res5 layers are employed in our MS-QAConv and MSDA-QAConv blocks. In DABN, we set the reduction rate $r = 16$.
We set the batch size $B = 64$, and we employ the graph sampler~\cite{DBLP:journals/corr/abs-2104-01546} to select 16 identities containing 4 samples in each ID. 
The input images are resized to 384$\times$128, and several data augmentation strategies, including random flipping, cropping, color jittering, are applied.
We adopt the SGD optimizer with a weight decay of 0.0005. The initial learning rate is set to 0.005 and is reduced by a factor of 10 after 20 epochs, with the training stage terminating at the 30-th epoch. All the experiments are implemented in PyTorch~\cite{DBLP:conf/nips/PaszkeGMLBCKLGA19}, with four NVIDIA 3090 GPUs.

%Our framework can be applied to both multi-source and single source DG tasks. In multi-source DG task, each source dataset can be naturally regraded as a domain. In single source DG, we split domains according to camera views, which display different statistics within a dataset.

\subsection{Datasets and Evaluation Protocols}
\textbf{Datasets.}\footnote{Note that DukeMTMC-reID~\cite{ristani2016performance} has been taken down and we did not employ it in our paper. DukeMTMC-reID is replaced by the RandPerson dataset \cite{wang2020surpassing} in our experiments.} We employ four large-scale datasets, CUHK03~\cite{li2014deepreid}, Market-1501~\cite{zheng2015scalable}, MSMT17~\cite{wei2018person}, and RandPerson~\cite{wang2020surpassing} in our experiments, among which RandPerson is a newly released synthesized person re-id dataset and the others are captured from real-world cameras. For CUHK03 dataset, we adopt CUHK03-NP~\cite{zhong2017re} detected subset which contains 767 subjects for training and 700 for evaluation. The widely used Market-1501 dataset includes 1,501 identities and 32,668 pedestrian images captured by 6 different cameras. MSMT17 is currently one of the largest public datasets obtained with real-world cameras. It consists of 4,101 annotated identities and 126,441 images from 12 outdoor and 3 indoor cameras. RandPerson is a synthesized re-id dataset generated by MakeHuman and Unity3D, and it contains a training set of 1,801,816 images and 8,000 synthesized identities. To make a fair comparison, we follow~\cite{DBLP:journals/corr/abs-2104-01546} to employ a subset of RandPerson, including 132,145 images of the 8,000 identities. Since RandPerson merely contains the training subset, it is not adopted as the target domain in our experiment.

\begin{table*}
\centering

% \begin{tabular}{c|c|c|c|c|c|c|c|c}
\begin{tabular}{p{2.8cm}<{\centering}|p{2.5cm}<{\centering}|p{2cm}<{\centering}|p{0.9cm}<{\centering}p{0.9cm}<{\centering} | p{0.9cm}<{\centering}p{0.9cm}<{\centering} | p{0.9cm}<{\centering}p{0.9cm}<{\centering}}
  \thickhline
  \rowcolor{mygray}  
  &  & {Training} & \multicolumn{2}{c|}{CUHK03-NP} & \multicolumn{2}{c|}{Market-1501} & \multicolumn{2}{c}{MSMT17} \\
  \cline{4-9}
  \rowcolor{mygray}  
  {\multirow{-2}{*}{Methods}}    & \multirow{-2}{*}{{Venue}} &  Data & top-1 & mAP & top-1 & mAP & top-1 & mAP \\
  \hline
  \hline
  MGN \cite{DBLP:conf/mm/WangYCLZ18} &	ACMMM 2018&	Market-1501&		8.5&	7.4&	-&	-&	-&	-\\
  MuDeep \cite{qian2019leader} & TPAMI 2020 & Market-1501 & 10.3 & 9.1 & - & - & - & - \\
  CBN \cite{zhuang2020rethinking} & ECCV 2020 & Market-1501  & - & - & - & - & 25.3 & 9.5 \\
  QAConv \cite{liao2020interpretable} & ECCV 2020 & Market-1501 & 9.9 & 8.6 & - & - & 22.6	&7.0 \\
 QAConv-GS \cite{DBLP:journals/corr/abs-2104-01546} &	arXiv 2021 &	Market-1501 	&16.4	&15.7	&-	&-	&41.2	&15.0\\
TAL (Ours)  &	- &	Market-1501 	&\textbf{20.2}	&\textbf{19.2}	&-	&-	& \textbf{43.5}	& \textbf{16.3} \\
\hline
\hline
PCB \cite{DBLP:conf/eccv/SunZYTW18}	&ECCV 2018&	MSMT17 &	-&	-&	52.7&	26.7&	-&	-\\
MGN \cite{DBLP:conf/mm/WangYCLZ18} &	ACMMM 2018&	MSMT17&	-&	-&	48.7&	25.1&	-&	-\\
ADIN \cite{yuan2020calibrated} &	WACV 2020&	MSMT17&		-&	-&	59.1&	30.3&	-&	-\\
SNR \cite{jin2020style}&	CVPR 2020&		MSMT17&		-&	-&	70.1&	41.4&	-&	-\\
CBN \cite{zhuang2020rethinking} &	ECCV 2020&	MSMT17&	-&	-&	73.7&	45.0&	-&	-\\
QAConv-GS \cite{DBLP:journals/corr/abs-2104-01546}	& arXiv 2021	& MSMT17	&20.0	&19.2	&75.1	&46.7	&-	&-\\
TAL (Ours)	& -	& MSMT17	&\textbf{20.4}	&\textbf{19.8}	&\textbf{77.5}	&\textbf{49.4}	&-	&-\\
\hline
\hline					
RP Baseline	\cite{wang2020surpassing} & ACMMM 2020	&RandPerson		&13.4	&10.8	&55.6	&28.8	&20.1	&6.3\\
QAConv-GS \cite{DBLP:journals/corr/abs-2104-01546}	&arXiv 2021	&RandPerson	&14.8	&13.4	&74.0	&43.8	&42.4	&14.4\\
TAL (Ours)	&-	&RandPerson	& \textbf{16.2}	& \textbf{14.2}	&\textbf{75.0}	&\textbf{45.9}	& \textbf{42.8}	& \textbf{14.9}\\
\hline
\end{tabular}
\caption{Comparison with the state-of-the-art methods in single-source DG re-id under protocol-1 .}\label{tab:sota1}
\end{table*}

\begin{table}
\centering
\begin{tabular}{c|cccc}
  \thickhline
  \rowcolor{mygray}
  Target: Market & mAP & top-1 & top-5 & top-10 \\
  \hline
  \hline
  QAConv-GS* \cite{DBLP:journals/corr/abs-2104-01546}& 59.3 & 83.0 &	92.8 &	95.0\\
  ${\rm M^3L}$* \cite{zhao2021learning} & 58.3 &	80.6 &	 91.2 &	94.2\\
    MetaBIN* \cite{DBLP:conf/cvpr/ChoiKJPK21} &57.9 & 80.3 & 91.2 & 93.7\\
  TAL (Ours) & \textbf{64.4} & \textbf{85.1}	&	\textbf{94.1}&	\textbf{96.2}\\ \hline
   \rowcolor{mygray}
  Target: CUHK03 & mAP & top-1 & top-5 & top-10 \\
  \hline
  \hline
  QAConv-GS* \cite{DBLP:journals/corr/abs-2104-01546}& 25.8 & 26.6	&	45.4 &	56.1\\
  ${\rm M^3L}$* \cite{zhao2021learning} & 33.9 &	34.8 &	55.4 &	66.3\\
    MetaBIN* \cite{DBLP:conf/cvpr/ChoiKJPK21}& 34.4 & 35.3 & 54.1 & 64.4\\
  TAL (Ours) & \textbf{35.0} &	\textbf{36.1} &	 \textbf{58.1} &	\textbf{66.8}\\ \hline
   \rowcolor{mygray}
  Target: MSMT17 & mAP & top-1 & top-5 & top-10 \\
  \hline
  \hline 
  QAConv-GS* \cite{DBLP:journals/corr/abs-2104-01546}& 20.2 & 52.1	&	64.4 &	69.2\\
  ${\rm M^3L}$* \cite{zhao2021learning} & 14.6 & 35.4 & 49.0	&	54.8\\
    MetaBIN* \cite{DBLP:conf/cvpr/ChoiKJPK21} & 16.3 & 40.3 & 53.9 & 59.5\\
  TAL (Ours) & \textbf{22.3} &	 \textbf{54.7} & \textbf{65.6}	&	\textbf{70.4}\\
 
\hline
\end{tabular}
\caption{Comparison with the state-of-the-art methods in multi-source DG re-id under protocol-2. * denotes the results are re-implemented based on the authors' code with the same source datasets.}\label{tab:sota2}
\end{table}

\textbf{Evaluation Protocols}. To fully verify the effectiveness of our two-stream adaptive learning framework, we design experiments under both single-source and multi-source protocols: (1) \emph{Protocol-1: single-source DG re-id.} Under this protocol , we perform training on the training set of one dataset or its subset, and evaluate on the test set of another dataset. As a single dataset does not naturally contain different domains, we split domains according to camera views, which display different statistics within a dataset. (2) \emph{Protocol-2: multi-source DG re-id.} Under this protocol, we employ RandPerson and two of the three real-world datasets (CUHK03, Market-1501 and MSMT17) as the training set, and the remaining one is used for evaluation. In this case, each source dataset can be naturally regarded as a domain.
For both settings, we adopt mean average precision (mAP) and CMC top-1, top-5, top-10 accuracy as the evaluation metrics.

\subsection{Comparison with State-of-the-arts}
% We compare the proposed TAL framework with the state-of-the arts DG re-id methods under both settings. The comparison on single-source DG re-id (protocol-1) is reported in Table~\ref{tab:sota1} and the multi-source DG re-id (protocol-2) results are shown in Table~\ref{tab:sota2}.

\textbf{Comparison Under Protocol-1}. We compare our TAL framework with several DG re-id methods, including MGN~\cite{DBLP:conf/mm/WangYCLZ18}, MuDeep~\cite{qian2019leader}, CBN~\cite{zhuang2020rethinking}, PCB~\cite{DBLP:conf/eccv/SunZYTW18}, ADIN~\cite{yuan2020calibrated}, SNR~\cite{jin2020style}, QAConv~\cite{liao2020interpretable} and QAConv-GS~\cite{DBLP:journals/corr/abs-2104-01546}. 
We successively employ Market-1501, MSMT17, and RandPerson as training sources, while the other real-world datasets are used for testing. As shown in Table~\ref{tab:sota1}, our proposed TAL clearly outperforms the state-of-the-art models on all three target datasets. TAL further improves our baseline models, \ie, QAConv and QAconv-GS. On common real-world~$\to$~real-world tasks, we achieve 0.6\% to 3.5\% improvements in terms of mAP and up to 3.8\% improvements for top-1 accuracy. The improvements on the more challenging synthesized~$\to$~real-world task are up to 1.4\% for top-1 and 2.1\% for mAP. Notably, our experiment confirms the finding in~\cite{wang2020surpassing}, which claims that models trained on a synthesized dataset could achieve competitive generalization ability as those trained on a real dataset.

\textbf{Comparison Under Protocol-2}. Under the multi-source protocol, we respectively adopt Market-1501, CUHK03, and MSMT17 as the test set, while all the other datasets are combined for training. To conduct fair comparisons, we use the authors' code to re-implement QAConv-GS~\cite{DBLP:journals/corr/abs-2104-01546}, $\rm M^3L$~\cite{zhao2021learning} and MetaBIN~\cite{DBLP:conf/cvpr/ChoiKJPK21} with the same training data as our method. From the comparisons shown in Table~\ref{tab:sota2}, we can make the following observations. First, our proposed TAL framework improves current methods by a large margin. For example, mAP is improved by 5.1\%, and top-1 is increased by 2.1\% on Market-1501. Second, state-of-the-arts multi-source DG re-id methods appear to be not stable, as QAConv-GS shows advantages on Market-1501 and MSMT17, but clear poor performance for both mAP and top-1 on CUHK03. In contrast, benefiting from the two-stream design for capturing both domain-specific and domain-invariant information, our method demonstrates notable superiority on all three test datasets.

\begin{table}[t]
\setlength{\abovecaptionskip}{2mm}
\centering
% \begin{tabular}{p{1.2cm}<{\centering}p{1.1cm}<{\centering}p{1.0cm}<{\centering}|p{1.5cm}<{\centering}p{1.5cm}<{\centering}}
\begin{tabular}{c|cc|cc|cc}
\hline\thickhline
\rowcolor{mygray}  
              &       &         & \multicolumn{2}{c|}{Target: M}    & \multicolumn{2}{c}{Target: C}                        \\ \cline{4-7} 
\rowcolor{mygray}  
 {\multirow{-2}{*}{backbone}}                  &  {\multirow{-2}{*}{DS}}                  &  {\multirow{-2}{*}{DI}}                & mAP                  & \multicolumn{1}{c|}{top-1}  & mAP                  & \multicolumn{1}{c}{top-1} \\ 
\hline \hline
 & $\times$   &$\times$ & 56.4  & 81.1     & 22.7 & 23.5  \\
 &$\checkmark$    & $\times$ & 57.5  & 82.0 & 25.4& 27.0                 \\
 & $\times$  &$\checkmark$    &  58.4  & 82.3 & 27.6 & 28.9                  \\
{\multirow{-3}{*}{ResNet-50}}    & $\checkmark$  & $\checkmark$  &\textbf{60.7}   & \textbf{82.7} & \textbf{30.0} & \textbf{31.6}           \\ \hline\hline
 &  $\times$  &$\times$ &  59.3 & 83.0     & 25.8 & 26.6   \\
& $\checkmark$  & $\times$& 60.1  & 83.1    &  31.0 & 33.6               \\
 &$\times$ & $\checkmark$& 62.4 & 83.8    & 32.7 & 34.9                         \\
{\multirow{-3}{*}{IBN-Net50}} & $\checkmark$  & $\checkmark$ & \textbf{64.4}  &  \textbf{85.1}  & \textbf{35.0} & \textbf{36.1}           \\\hline
\end{tabular}
\caption{Comparative results when employing different backbones with different network branches. M: Market-1501. C: CUHK03.}
\label{tab:ts}
\end{table}

\begin{table}[t]
\setlength{\abovecaptionskip}{2mm}
\centering
% \begin{tabular}{p{1.2cm}<{\centering}p{1.1cm}<{\centering}p{1.0cm}<{\centering}|p{1.5cm}<{\centering}p{1.5cm}<{\centering}}
% \begin{tabular}{c|cc|cc}
\begin{tabular}{p{2.5cm}<{\centering}|p{0.9cm}<{\centering}p{0.9cm}<{\centering}|p{0.9cm}<{\centering}p{0.9cm}<{\centering}}
\hline\thickhline
\rowcolor{mygray}  
              &     \multicolumn{2}{c|}{Target: M}    & \multicolumn{2}{c}{Target: C}                        \\ \cline{2-5} 
\rowcolor{mygray}  
 {\multirow{-2}{*}{DI Network}}            & mAP                  & \multicolumn{1}{c|}{top-1}  & mAP                  & \multicolumn{1}{c}{top-1} \\ 
\hline \hline  
Average Pooling &  61.5  &  83.4    & 31.5 & 34.0                \\
BN & 61.0    & 83.2   & 31.9 & 34.3               \\
DABN   & \textbf{62.4} & \textbf{83.8}    & \textbf{32.7} & \textbf{34.9}      \\         \hline
\end{tabular}
\caption{Comparative results when employing different training strategies in the DI network. M: Market-1501. C: CUHK03.}
\label{tab:dabn}
\end{table}

\subsection{Ablation Study}
In this section, we analyze the effectiveness of our TAL framework and conduct comprehensive ablation studies on the detailed components of TAL. We report the results under protocol-2 with the target datasets Market-1501 (M) and CUHK03 (C).

\textbf{Effectiveness of Two-stream Modeling}. Since TAL aims to simultaneously capture the domain-specific and domain-invariant information in a single framework, it is important to investigate how each sub-network influences the overall performance and how they work together. To this end, we perform independent evaluations on the output of each network, and the results are reported in Table~\ref{tab:ts}. With neither the DS nor the DI networks, our framework degenerates into QAConv-GS~\cite{DBLP:journals/corr/abs-2104-01546}, which could be regarded as our baseline. By introducing the DS and DI networks, the performances are continuously improved. Finally, by combining the results of both the DS and DI networks under the IBN-Net50 backbone, the performance of TAL is further improved, \ie, improving the baseline by 5.1\% and 9.2\% in mAP on Market-1501 and CUHK03, respectively. In the meantime, significant improvements can also be achieved with the ResNet-50 backbone. These results indicate that both the DS and DI information plays an important role in enhancing the generalization ability of the re-id model, and in TAL, the two-stream structure successfully captures both clues as complements to each other.

%To fully investigate the necessity of each branch of our two-stream modeling design, we conduct ablation studies with different backbones in Table~\ref{tab:ts}. When using IBN-Net50 as backbone, it can be observed that mAP increases 4-4.3\% and Rank-1 increases 2.5-2.8\% with DI branch. In the same way, DS branch improves the performance by 2-2.3\% for mAP and 1.2-1.3\% for Rank-1. Similar results can be observed for ResNet-50 backbone. These observations validate the hypothesis that both DS and DI statistics are indispensible to generalization ability of re-id model, and our two-stream framework successfully enhances both of them.

\textbf{Effectiveness of Domain Adaptive BN}. To evaluate the effectiveness of DABN in adaptively aggregating domain-specific information, we compare it with an average pooling and a simple BN baseline. From Table~\ref{tab:dabn}, it can be observed that by replacing the attention mechanism with an average pooling operation, the performance is decreased by $\sim$1\% on both datasets. In the meantime, replacing DABN with a vanilla BN layer also deteriorates the performance. These results indicate that DABN could facilitate the fusion of statistics from the domain-specific BNs and effectively approximate the unseen target domain information.

\begin{table}[t]
\setlength{\abovecaptionskip}{2mm}
\centering
% \begin{tabular}{p{1.2cm}<{\centering}p{1.1cm}<{\centering}p{1.0cm}<{\centering}|p{1.5cm}<{\centering}p{1.5cm}<{\centering}}
% \begin{tabular}{c|cc|cc}
\begin{tabular}{p{2.5cm}<{\centering}|p{0.9cm}<{\centering}p{0.9cm}<{\centering}|p{0.9cm}<{\centering}p{0.9cm}<{\centering}}
\hline\thickhline
\rowcolor{mygray}  
              &     \multicolumn{2}{c|}{Target: M}    & \multicolumn{2}{c}{Target: C}                        \\ \cline{4-5} 
\rowcolor{mygray}  
 {\multirow{-2}{*}{DS network}}            & mAP                  & \multicolumn{1}{c|}{top-1}  & mAP                  & \multicolumn{1}{c}{top-1} \\ 
\hline \hline  
Average Pooling &  58.2  & 82.5   & 29.4 & 31.1                 \\
Voting & 58.5  & 82.4      & 30.1  & 31.7                \\
MSDA-QAConv  & \textbf{60.1}  & \textbf{83.1}    &  \textbf{31.0} & \textbf{33.6}          \\         \hline
\end{tabular}
\caption{Comparative results when employing different training strategies in the DS network. M: Market-1501. C: CUHK03.}
\label{tab:dam}
\end{table}

\begin{table}[t]
\setlength{\abovecaptionskip}{2mm}
\centering
% \begin{tabular}{p{1.2cm}<{\centering}p{1.1cm}<{\centering}p{1.0cm}<{\centering}|p{1.5cm}<{\centering}p{1.5cm}<{\centering}}
\begin{tabular}{p{1cm}<{\centering}p{1cm}<{\centering}|p{0.9cm}<{\centering}p{0.9cm}<{\centering}|p{0.9cm}<{\centering}p{0.9cm}<{\centering}}
% \begin{tabular}{cc|cc|cc}
\hline\thickhline
\rowcolor{mygray}  
     &         & \multicolumn{2}{c|}{Target: M}    & \multicolumn{2}{c}{Target: C}                        \\ \cline{3-6} 
\rowcolor{mygray}  
  {\multirow{-2}{*}{res4}}                  &  {\multirow{-2}{*}{res5}}                & mAP                  & \multicolumn{1}{c|}{top-1}  & mAP                  & \multicolumn{1}{c}{top-1} \\ 
\hline \hline  
  $\checkmark$ & $\times$ & 61.7 & 83.2  &   28.9    &  30.6                 \\
   $\times$  & $\checkmark$ & 62.5 &    83.9     & 30.4  & 32.2                    \\
 $\checkmark$  & $\checkmark$ &  \textbf{64.4}  & \textbf{85.1}    & \textbf{35.0} & \textbf{36.1}          \\\hline
\end{tabular}
\caption{Comparative results when employing different levels of features in backbone features. M: Market-1501. C: CUHK03.}
\label{tab:ms}
\end{table}

\textbf{Effectiveness of Domain Adaptive Matching}. In the DS network, we employ MSDA-QAConv to aggregate the correspondence maps and then perform matching. To evaluate its contribution, we compare it with two baselines. First, instead of domain-adaptive attention, we employ the average response map to calculate the matching results. As shown in Table~\ref{tab:dam}, this causes 1.6\% to 1.9\% performance drop in mAP. Second, we remove the MSDA-QAConv block, and directly sum up the voting results from domains-specific matchers. This also results in inferior performance. These results verify the strong adaptability of the proposed domain-adaptive matching strategy.

%Table~\ref{tab:dam} reports the ablation results of different training strategies in MoE. It can be observed that $conclusion$. Our proposed MSDA-QAConv significantly outperforms the explicit voting method for its better feature aggregation ability. Specifically, $reasons$.

\textbf{Effectiveness of Multi-Scale Matching}. We conduct experiments by employing different levels of features in the backbone for calculating the similarity between image pairs. Specifically, we only investigate the features from the res4 and res5 layers, because low-level features from res2 and res3 layers are with relatively high resolution and will cause memory and efficiency issues.  
As shown in Table~\ref{tab:ms}, although single-level features also yield promising results, combining them further significantly improves the performance. These results demonstrate that multi-scale features not only work well in supervised re-id tasks, but also benefit DG re-id.

\begin{figure}
    \centering
    \includegraphics[width=\linewidth]{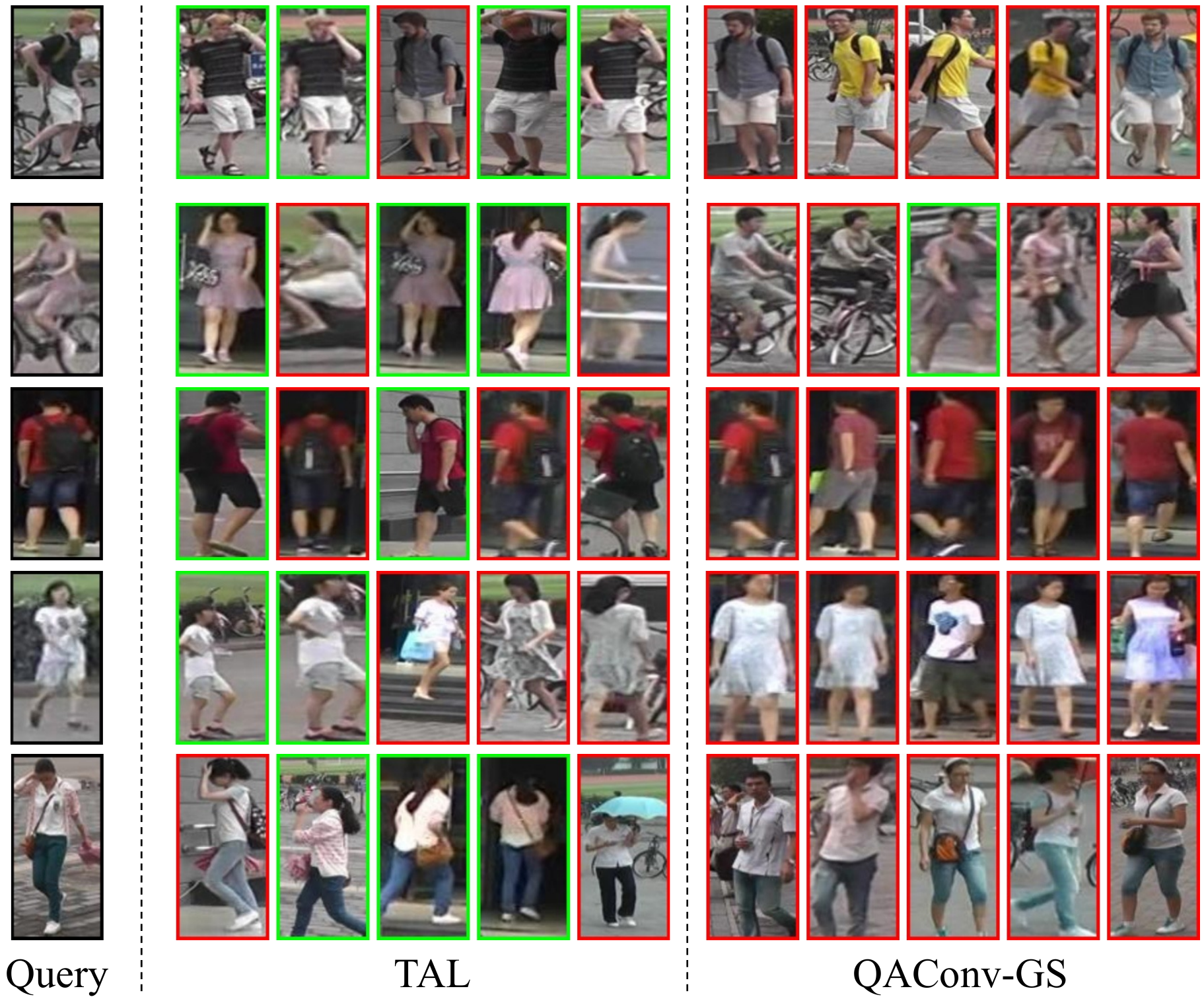}
    \caption{Visualization of the top-5 ranking results. These queries are difficult for the QAConv-GS \cite{DBLP:journals/corr/abs-2104-01546} model to retrieve, but much better results are achieved by TAL. The green and red bounding boxes denote the correct and incorrect matches, respectively.}
    \label{fig:qr1}
\end{figure}

\subsection{Qualitative Results}
To better illustrate the improvements achieved with our framework, we visualize some qualitative matching results in Fig.~\ref{fig:qr1}. We can observe that our model (TAL) successfully retrieves the difficult queries from the target domain under various challenging conditions, including occlusion, pose/viewpoint variation, where the baseline model (QAConv-GS) fails. Notably, although QAConv-GS also adopts a local matching strategy, it fails to accurately differentiate persons with similar appearances. In contrast, our model performs much better under this circumstance. We believe this is because our framework simultaneously captures DS and DI information, which makes TAL more robust to the feature variation caused by domain shift.

\begin{figure}
    \centering
    \includegraphics[width=\linewidth]{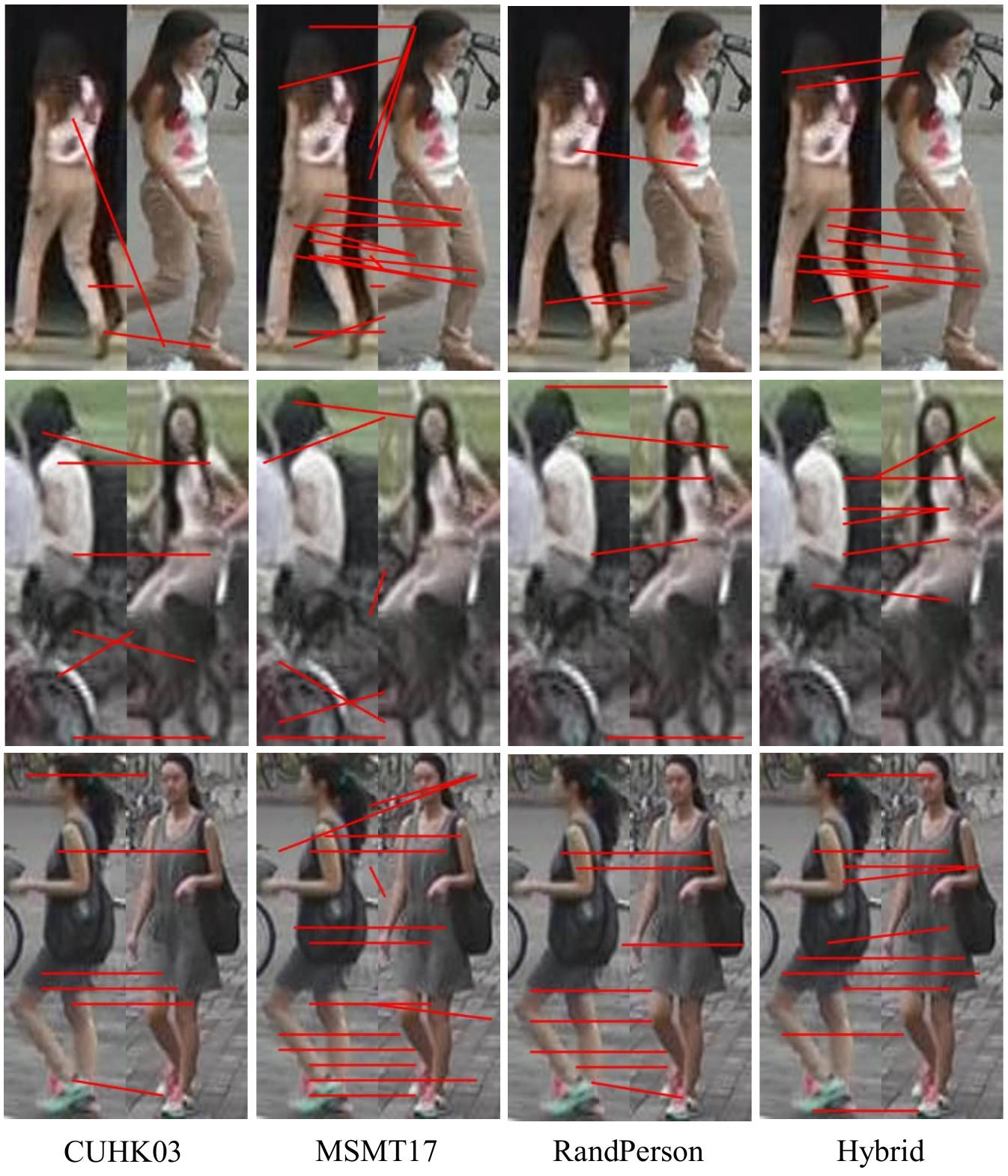}
    \caption{Examples of the correspondences between local patches. We evaluate on Market-1501 while the model is trained on CUHK03, MSMT17, and RandPerson. The left three columns show the matching results generated by three domain experts in the DS network, and the right column shows the results generated by the DI network with hybrid inputs. }
    \label{fig:qr2}
\end{figure}

To better present the working mechanism of the DS and DI networks, we visualize the local correspondences between the query and gallery images. As shown in Fig.~\ref{fig:qr2}, the domain experts in the DS network generate different matching patterns. For example, the expert trained on CUHK03 focuses more on the head and leg regions, maybe because these regions in Market-1501 have similar statistics with CUHK03. In contrast, the correspondences generated by the DI network trained with the hybrid dataset tend to locate in more diverse regions, \eg, head, body, and leg, which are critical for identifying a person. This demonstrates that the DS and DI networks provide complementary information to each other to better identify a person in unseen domains.

\section{Conclusion and Limitation}
In this paper, we propose a two-stream framework to address the challenges in DG re-id, where the domain-specific and domain-invariant features are simultaneously captured in a unified framework. We also design two adaptive learning modules to facilitate domain generalizable learning. Extensive results on both single-source and multi-source DG re-id benchmarks demonstrate that the proposed framework significantly outperforms the state-of-the-art methods.

\textbf{Limitation.} Although the employment of DSBN makes our framework more efficient than traditional MoE frameworks, it still faces challenges in terms of memory and speed when there are a large number of training domains, \eg, 15 camera views on the MSMT17 dataset. In this work, we replace all the BN layers with DSBN. It would be interesting to explore whether some BN layers can be shared, or introduce a dynamic learning strategy~\cite{DBLP:conf/cvpr/LiYCWV21} in DSBN. This will further improve the efficiency of our framework.

%%%%%%%%% REFERENCES
{\small
\bibliographystyle{ieee_fullname}
\bibliography{egbib}
}

\end{document}